\documentclass[twoside,11pt]{article}

\usepackage{jmlr2e}
\usepackage{amsmath}
\usepackage{caption}
\usepackage{subcaption}
\usepackage{algorithm}
\usepackage{algpseudocode}
\algnewcommand{\algorithmicand}{\textbf{ and }}
\algnewcommand{\algorithmicor}{\textbf{ or }}
\algnewcommand{\OR}{\algorithmicor}
\algnewcommand{\AND}{\algorithmicand}
\algnewcommand{\var}{\texttt}

\jmlrheading{1}{}{}{}{}{}

\firstpageno{1}

\begin{document}

\title{Online Active Learning for Soft Sensor Development using Semi-Supervised Autoencoders}

\author{\name Davide Cacciarelli \email dcac@dtu.dk\\
       \name Murat Kulahci \email muku@dtu.dk\\
       \addr Department of Applied Mathematics and Computer Science\\
       Technical University of Denmark\\
       Kgs. Lyngby, Denmark
       \AND
       \name John Tyssedal \email john.tyssedal@ntnu.no \\
       \addr Department of Mathematical Sciences\\
       Norwegian University of Science and Technology\\
       Trondheim, Norway}

\maketitle

\begin{abstract}%   <- trailing '%' for backward compatibility of .sty file
Data-driven soft sensors are extensively used in industrial and chemical processes to predict hard-to-measure process variables whose real value is difficult to track during routine operations. The regression models used by these sensors often require a large number of labeled examples, yet obtaining the label information can be very expensive given the high time and cost required by quality inspections. In this context, active learning methods can be highly beneficial as they can suggest the most informative labels to query. However, most of the active learning strategies proposed for regression focus on the offline setting. In this work, we adapt some of these approaches to the stream-based scenario and show how they can be used to select the most informative data points. We also demonstrate how to use a semi-supervised architecture based on orthogonal autoencoders to learn salient features in a lower dimensional space. The Tennessee Eastman Process is used to compare the predictive performance of the proposed approaches.
\end{abstract}


\section{Introduction}
In industrial operations, soft sensors are frequently used for real-time prediction of hard-to-measure process variables, as well as to support system backup strategies, what-if analysis, sensor validation, and fault diagnosis \citep{Fortuna2007}. Soft sensors are classified into two types: model-driven sensors, which are used to depict the ideal steady-state of a process under normal operating conditions, and data-driven sensors, which are used to better approximate real process conditions \citep{KADLEC2009795}. Many labeled observations are required for training the regression models used in soft sensor development, but in industrial contexts, data is often abundant only in an unlabeled form. Obtaining product information can be both expensive and time consuming, as it may necessitate the intervention of a human expert or the use of expensive testing equipment. As a result, active learning is becoming increasingly useful for reducing the number of labels required to achieve compelling predictive performance.
Active learning-based sampling schemes use some evaluation criteria to assess the informativeness of the unlabeled data points and prioritize the labeling of the most useful instances for building the model. Three macro scenarios can be identified depending on how the unlabeled instances are fed into the learner and then selected to be labeled by an oracle \citep{Settles2009}. The first scenario is referred to as membership query synthesis, and it allows the learner to query the labels of synthetically generated instances rather than those sampled from the process distribution. The second scenario is stream-based active learning, also known as selective sampling. It denotes a situation in which instances are drawn sequentially and the learner must immediately decide whether to keep the instance and query its label or discard it. The third and final scenario is pool-based active learning, which depicts a situation where a large amount of unlabeled data is collected all at once and made available to the learner, which can rank all of the data points and select the most informative ones.  While many researchers have been working on active learning in the latest years, pool-based active learning for classification has received the most attention \citep{Cai2013}. 

In this work, we focus on stream-based active learning \citep{surveyOAL}, which represents a more difficult task as the learner cannot observe all of the available observations before deciding which labels to query. We believe that this scenario accurately reflects high-volume production processes in which samples are processed very rapidly and labels are no longer retrievable. Stream-based active learning should be considered and prioritized for all industrial processes with similar properties.

\section{Background}
In regression modeling, we try to learn a function $\hat{f}: \boldsymbol{x} \in \mathbb{R}^p \rightarrow y \in \mathbb{R}$ to predict a quality characteristic or a hard-to-measure variable $y \in \mathbb{R}$ that is related to other process variables $\boldsymbol{x} \in \mathbb{R}^p$. Accordingly with many active learning approaches \citep{Cai2013}, we assume a labeled dataset $\mathcal{L} = \{(\boldsymbol{x}_i, y_i)_{i=1}^n, \boldsymbol{x}_i \in \mathbb{R}^p, y \in \mathbb{R}\}$ with $n$ observations is initially available to fit a linear regression model of the kind
\begin{equation*}
    f(\boldsymbol{x}; \boldsymbol{\beta}) = \sum_{i=0}^{p}\beta_i x_i = \boldsymbol{\beta}^T\boldsymbol{x}
\end{equation*}
where $x_0=1$ is the intercept term and $x_i$ with $i=1, ..., p$ are the $p$ process variables. Parameters are estimated by minimizing a squared error loss given by
\begin{equation}
\label{eqn:error}
    \hat{\epsilon} = \frac{1}{n}\sum_{i=1}^{n}(y_i - f(\boldsymbol{x}_i))^2
\end{equation}
After an initial model has been built, we aim to acquire additional observations by evaluating the unlabeled data points, until a budget constraint is met. Some commonly encountered approaches are presented below.

\subsection{Mahalanobis Distance}
If we only examine the feature space, a desirable property that we might pursue when collecting instances for our training set $\mathcal{L}$, is to ensure diversity among the observations. The Hotelling $T^2$ control chart, which is widely used in statistical process control (SPC) to detect anomalous data points \citep{Hotelling1947}, can accomplish so. As we do in SPC, in this case we use Mahalanobis distance (Equation \ref{eqn:hotelling}) to measure the dissimilarity between the new unlabeled instances and the observations in the current training set $\mathcal{L}$. The Hotelling $T^2$ statistic for a new unlabeled instance $\boldsymbol{x}$ is computed as
\begin{equation}
\label{eqn:hotelling}
    T^2(\boldsymbol{x}) = (\boldsymbol{x}-\bar{\boldsymbol{x}})^T\boldsymbol{S}^{-1}(\boldsymbol{x}-\bar{\boldsymbol{x}})
\end{equation}
where $\bar{\boldsymbol{x}}$ and $\boldsymbol{S}$ correspond to the sample mean vector and sample covariance matrix of $\mathcal{L}$, respectively. This approach has been extended to a principal component regression (PCR) model by \cite{Ge2014}, who proposed a sampling index dependent on the Hotelling $T^2$ statistic and the squared prediction error. In this case, the sampling function is simply represented by $argmax_{\boldsymbol{x}} T^2(\boldsymbol{x})$.

\subsection{Query By Committee}
While the previous approach only considers the feature space, query by committee (QBC) tries to evaluate the uncertainty about the response. This strategy, initially introduced for classification problems, was extended to regression tasks by \cite{Burbidge}. The main intuition is that by building an ensemble of regression models trained on bootstrap replica of the original training set $B(K)=\{f_1,f_2,..., f_K\}$ we can approximate the  distribution of the predictive variance. Once the ensemble has been built, we can measure the variance of the predictions made by the committee members for each unlabeled observation $\boldsymbol{x}$. This variance, or ambiguity, is computed as
\begin{equation}
\label{eqn:ambiguity}
    a(\boldsymbol{x}) = \frac{1}{K}\sum_{i=1}^{K}(f_i(\boldsymbol{x})-y_K(\boldsymbol{x}))^2
\end{equation}
where $y_K(\boldsymbol{x})$ is the mean of the predictions made by the ensemble members. The sampling function then simply becomes $argmax_{\boldsymbol{x}} a(\boldsymbol{x})$. The key intuition is that if many models disagree on the label associated with an instance, that instance is an ambiguous one.

\subsection{Expected Model Change}
Introduced by \cite{Cai2013}, expected model change (EMC) suggests querying the unlabeled example that would cause the maximum change in the current model parameters, if we knew its label. The model change is measured as the difference between the current model parameters and the parameters obtained after fitting the model on the enlarged training set $\mathcal{L}^+ = \mathcal{L} \cup (\boldsymbol{x}^+, y^+)$. The gradient of the loss is used to estimate the model change. Considering the augmented training set $\mathcal{L}^+$, the loss function shown in Equation \ref{eqn:error} becomes
\begin{equation*}
% \label{eqn:error}
    \hat{\epsilon} = \frac{1}{n}\sum_{i=1}^{n}(y_i - f(\boldsymbol{x}_i))^2 + (y^+ - f(\boldsymbol{x}^+))^2
\end{equation*}
where the last term, which is hereinafter referred to as $\ell_{\boldsymbol{x}^+}(\boldsymbol{\beta})$, represents the difference between the loss measured with the model trained on $\mathcal{L}$ and the one trained on $\mathcal{L}^+$. The derivative of the marginal loss $\ell_{\boldsymbol{x}^+}(\boldsymbol{\beta})$ with respect to the parameters $\boldsymbol{\beta}$ in the new point $\boldsymbol{x}^+$ is given by
\begin{equation}
\label{eqn:newerror}
    \begin{aligned}
    \hat{\epsilon} = \frac{\partial \ell_{\boldsymbol{x}^+}(\boldsymbol{\beta})}{\partial \boldsymbol{\beta}}
    & = 2(y^+ - f(\boldsymbol{x}^+)) \frac{\partial f(\boldsymbol{x}^+)}{\partial \boldsymbol{\beta}}
    \\ & = 2(y^+ - f(\boldsymbol{x}^+)) \frac{\partial \boldsymbol{\beta}^T\boldsymbol{x}^+}{\partial \boldsymbol{\beta}}
    \\ & = 2(y^+ - f(\boldsymbol{x}^+))\boldsymbol{x}^+
    \end{aligned}
\end{equation}
Since we do not know the true label of $\boldsymbol{x}^+$,  $y^+$ it is going to be replaced by the predictions $f_i(\boldsymbol{x}^+)$ made by the members of the bootstrap ensemble $B(K)$. 
Finally, the sampling function is given by $argmax_{\boldsymbol{x}} \frac{1}{K} \sum_{i=1}^{K} \Vert(f_i(\boldsymbol{x}) - f(\boldsymbol{x}))\boldsymbol{x} \Vert$.

\section{Proposed Approach}
Given the impossibility of ranking unlabeled instances in real-time and deterministically optimizing the sampling criteria, we propose leveraging unlabeled data to impose a threshold, or upper control limit (UCL), on the informativeness of the incoming data points. The unlabeled data pool can be acquired by either observing the process for a period of time without sampling the product information $y$ or by using data that is already available in the form of a historical database $\mathcal{H} = \{(\boldsymbol{x}_i), \boldsymbol{x}_i \in \mathbb{R}^p\}$. The primary difference between pool-based active learning and online active learning is that the labels of observations pertaining to $\mathcal{H}$ can no longer be queried because they only exist digitally, and the associated physical part or component is no longer available. The data in $\mathcal{H}$ is used to estimate the distribution of the statistics employed by the criteria in Equations \ref{eqn:hotelling}, \ref{eqn:ambiguity}, and \ref{eqn:newerror}. In this study, we employed kernel density estimation with a Gaussian kernel. The UCL is then determined by specifying the appropriate sampling rate $\alpha$. For a given criterion $\mathcal{J}$,  the threshold is defined as
\begin{equation}
\label{eqn:threshold}
    P(\mathcal{J(\boldsymbol{x})} \geq UCL) = \alpha
\end{equation}
Using the UCL obtained from Equation \ref{eqn:threshold}, we should then only collect the $\alpha$-percent most informative data points, according to the specific criterion $\mathcal{J}$ \citep{CACCIARELLI2022109664, Cacciarelli2023}. In this work, we test the stream-based active learning routine using Mahalanobis distance, ambiguity, and expected model change as sampling criteria $\mathcal{J}$. Before starting the active learning routine and collecting additional observations, we also propose to use a semi-supervised architecture by training an autoencoder network on the historical data $\mathcal{H}$. With semi-supervised learning, we can exploit all the available unlabeled data and learn how to extract relevant features that could be better predictors than the raw input features. Indeed, when variables are highly correlated, it has been demonstrated in the literature that a PCR model can be enhanced with semi-supervision \citep{Frumosu2018}. With regards to deep learning methods, autoencoders have been proposed to deal with semi-supervised learning in fault classification \citep{Jia2020, Jiang2017}. Recently, autoencoders have also been investigated in soft sensors applications \citep{Yuan2018, Moreira2021} but their contribution to the stream-based active learning scenario has not been evaluated yet. In this work, we propose the use of a semi-supervised architecture as the one shown in Figure 1. An orthogonal autoencoder (OAE) is employed for feature extraction and the encoded features are then used as predictors in a linear regression model. An OAE is an autoencoder network that minimizes an Ortho-Loss, which is comprised of a squared reconstruction loss and an orthogonality regularization term \citep{Wang2019, CACCIARELLI}. The regularization, weighted by a parameter $\lambda$, encourages the network to learn uncorrelated features in its bottleneck. This is particularly beneficial to alleviate the multicollinearity issue in the regression modeling stage.

\begin{figure}[hbt!]
\centering
    \includegraphics[width=.6\textwidth]{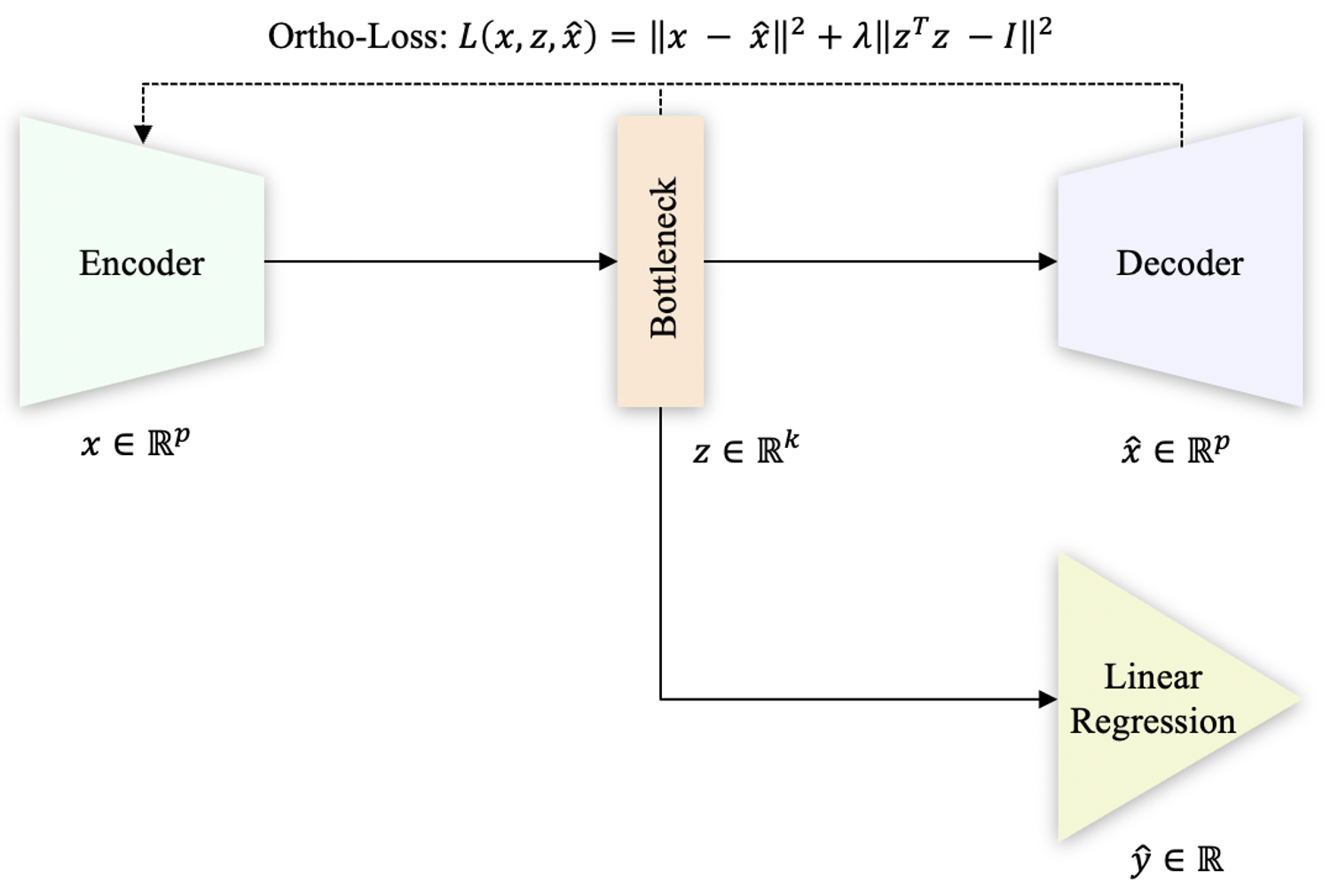}
\caption{Semi-supervised architecture based on OAE.}
\label{fig:ss-oae}       % Give a unique label
\end{figure}

The main advantage of the semi-supervised model is that the extracted features are more expressive than the original process variables. However, if the dimensionality of the bottleneck is lower than the one of the input features, there is an additional benefit for active learning. Indeed, because the majority of the provided active learning approaches are model-based, an initial number of labels is required. QBC and EMC, in particular, employ the linear regression model’s predictions to select the data points that should be queried. To uniquely determine the coefficients of a regression model, we need a number of observations larger than the number of parameters $\boldsymbol{\beta}$ to be estimated. This initial set of observations is usually collected at random \citep{Cai2013}. As a result, by reducing the dimensionality of the parameters $\boldsymbol{\beta}$, we will be able to anticipate the active collection phase and get more robust estimates for the same experimental cost. The complete active learning routine is reported Appendix A.

\section{Experiments}
The Tennessee Eastman Process (TEP) is considered the gold-standard benchmark for testing process control approaches \citep{Lawrence, Capaci2019} and, recently, it has also been used for validating active learning and soft sensor development methods \citep{Zhu2015, ratko}.
%For this study, we generated 50 datasets using the MATLAB code provided by \cite{Reinartz2021} and \cite{Andersen2022}.
The variables that have been used as predictors in regression modeling are the  same 16 controlled process variables used by \cite{Zhu2015} and \cite{ratko}, and the continuous response is Stream 9E, a composition measurement belonging to the purge stream.

\begin{figure}[hbt!]
\centering
\begin{subfigure}{.48\textwidth}
  \centering
  \includegraphics[width=1\linewidth]{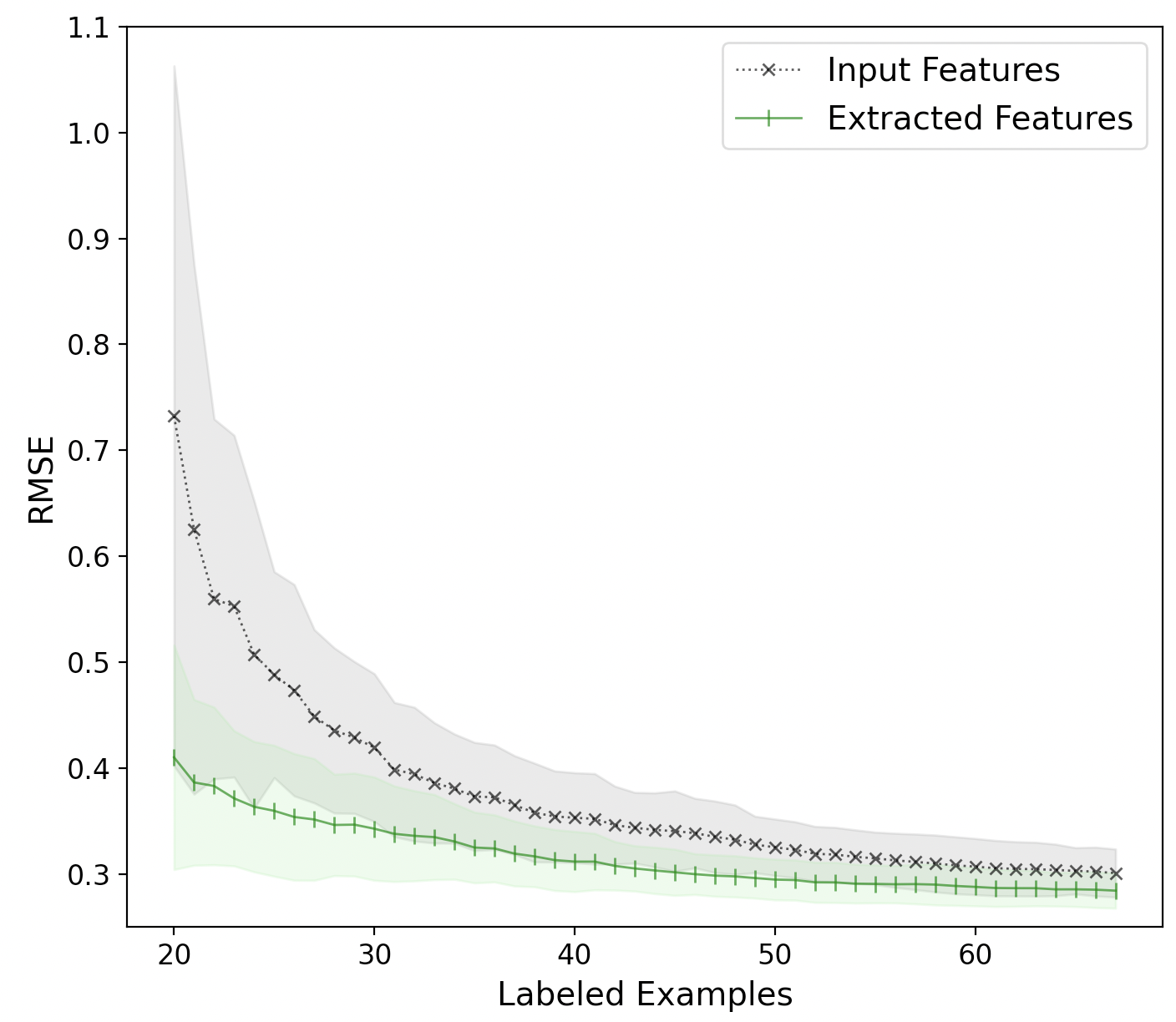}
  \caption{Semi-Supervised Learning}
  \label{fig:sub1}
\end{subfigure}%
\begin{subfigure}{.48\textwidth}
  \centering
  \includegraphics[width=1\linewidth]{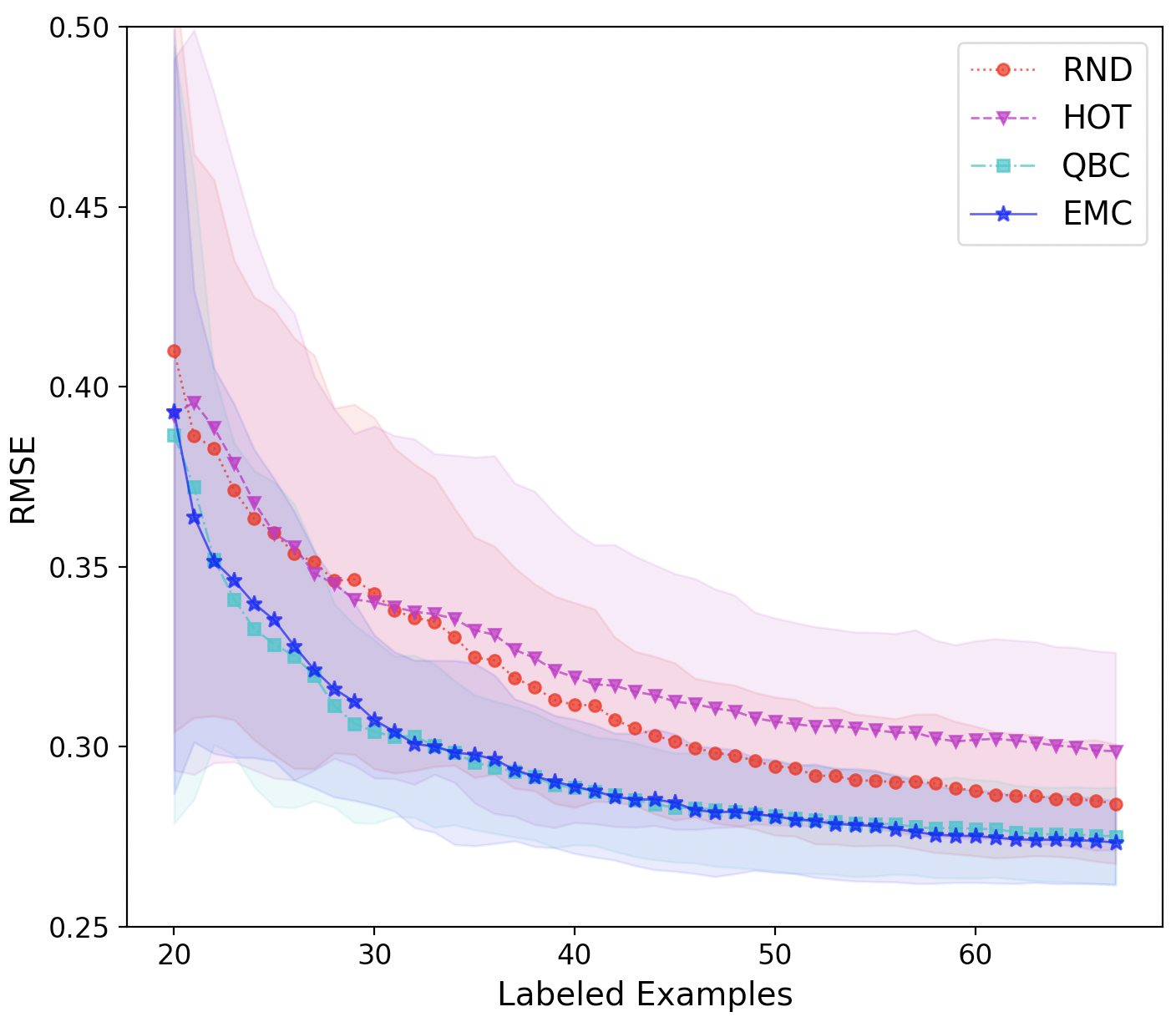}
  \caption{Active Learning}
  \label{fig:sub2}
\end{subfigure}
\caption{(a) shows the learning curves with random sampling using the original process variables and the features extracted by the OAE and (b) shows the learning curves of different active learning methods: random (RND), Hotelling $T^2$ (HOT), query by committee (QBC), and expected model change (EMC).}
\label{fig:test}
\end{figure}

Figure 2a shows how the suggested semi-supervised architecture can increase the predictive performance. Data is randomly sampled in both situations, but the two linear models are fitted using the original process variables and the features extracted by the OAE. 
%The OAE used in this study has an encoder with layers dimensionality equal to [16, 160, 80, 40, 20, 10], where 16 corresponds to the input dimension and 10 to the bottleneck; the decoder is symmetrically designed. 
To ensure comparability between the two learning curves, we fitted the first model when the number of gathered observations exceeded 16. We believe the improvement is due to the OAE's ability to express nonlinear relationships in data in its encoded features and to the fact that with the extracted features, we have the same number of observations to estimate a smaller number of parameters. In Figure 2b, we try to improve the semi-supervised result by using the proposed active learning strategies. It is clear that EMC and QBC consistently outperform the passive random approach in terms of recommending the most informative data points. On the contrary, the Mahalanobis distance appears to worsen the predictive performance. We believe this may due to the fact that data points with high $T^2$ statistics may be outliers, whose inclusion in the training set eventually degrades performance. It should be noted that in Figure 2b all the sampling strategies use the features extracted by the OAE. Experimental setup and training details are reported in Appendix B.

\section{Conclusion}
Industrial data is often only available unlabeled as quality inspections and manual annotation tasks are costly and time-consuming. In this work, we proposed a semi-supervised model based on OAEs for extracting relevant features and reducing multicollinearity. On top of this, we reviewed and adapted for the online setting some of the most widely used active learning strategies for linear regression. The analysis demonstrates how properly using the historical data and taking into account the expected response allows for a faster reduction of the prediction error. For future research, we will consider more advanced architectures such as LSTM autoencoders or transformers to obtain encoded features that take into account the temporal dependency in the data.

\newpage

\appendix
\section*{Appendix A. Online Active Learning Routine}
\begin{algorithm}
\caption{}\label{alg:1}
\begin{algorithmic}[1]
\Require a historical unlabeled dataset $\mathcal{H}$, a labeled dataset $\mathcal{L}$, a data stream $\mathcal{S}$, a budget $b$, and a criterion $\mathcal{J}$.
\State Train an OAE on $\mathcal{H}$
\State Encode observations in $\mathcal{H}$ and $\mathcal{L}$: $\boldsymbol{x} \in \mathbb{R}^p \longrightarrow \boldsymbol{z} \in \mathbb{R}^k$
\State Fit a linear regression model on the encoded features $\boldsymbol{z}$ and labels $y$ obtained from $\mathcal{L}$
\State Compute $\mathcal{J}$ on the encoded features $\boldsymbol{z}$ pertaining to $\mathcal{H}$ and estimate a threshold (UCL) using Equation 5
\State $i \gets 0$, $c \gets 0$
\While{$c \leq b$ \AND $i\leq |\mathcal{S}|$}
    \State Encode i\textit{th} observation from the stream $\mathcal{S}$: $\boldsymbol{x}_i \in \mathbb{R}^p \longrightarrow \boldsymbol{z}_i \in \mathbb{R}^k$
    \If{$\mathcal{J}(\boldsymbol{z}) \geq UCL$}
        \State Ask for the label $y_i$ and augment the labeled dataset: $\mathcal{L}^+ = \mathcal{L} \cup (\boldsymbol{z}_i, y_i)$
        \State $c \gets c+1$
        \State Update model (repeat Step 3)
        \State Update threshold (repeat Step 4)
    \Else
        \State Discard $\boldsymbol{x}_i$
    \EndIf
    \State $i \gets i+1$
\EndWhile
\end{algorithmic}
\end{algorithm}

\section*{Appendix B. Experimental Setup and Training Details}
The data is generated using the MATLAB code provided by \cite{Reinartz2021} and \cite{Andersen2022} with the Ricker closed-loop simulation model. No faults have been introduced throughout the 50 simulation runs, which are generated providing different seeds to the simulator. The variables used are reported in table \ref{table:1}. Sample rate was set to 1 minute.
\\\\
The active learning routine was tested once on each of the 50 simulation runs. Figure 2 reports the mean and standard deviation for each method across these 50 runs (shaded regions indicate $\pm$ 1 standard deviation). \\\\
With regards to the autoencoder structure, we used an encoder whose dimensionality of the layers corresponds to [16, 160, 80, 40, 20, 10]. The decoder is symmetrical to the encoder. The penalty term corresponding to the weight of the orthogonality regularization in the loss function was set to 0.10. No fixed number of epochs was used for the training as we followed an early stopping approach, setting a patience of 10 on the number of accepted epochs without improvement on the validation loss (20\% of the training data is used for validation). Finally, the bandwidth used for the kernel density estimation of the UCL is found using Scott's rule \citep{Scott}.

\begin{table}[hbt!]
\centering
\begin{tabular}{|| c | c ||}
\hline
\textbf{Process Variable}                       & \textbf{ID}   \\
 \hline\hline
A Feed (Stream 1)                               & XMEAS 1     \\
D Feed (Stream 2)                               & XMEAS 2      \\
E Feed (Stream 3)                               & XMEAS 3      \\
A and C Feed (Stream 4)                         & XMEAS 4      \\
Recycle Flow (Stream 8)                         & XMEAS 5      \\
Reactor Feed Rate (Stream 6)                    & XMEAS 6      \\
Reactor Temperature                             & XMEAS 9      \\
Purge Rate (Stream 9)                           & XMEAS 10      \\
Separator Temperature                           & XMEAS 11      \\
Separator Pressure                              & XMEAS 13      \\
Product Separator Underflow (Stream 10)         & XMEAS 14      \\
Stripper Pressure                               & XMEAS 16      \\
Stripper Temperature                            & XMEAS 18      \\
Stripper Steam Flow                             & XMEAS 19      \\
Reactor Cooling Water Outlet Temperature        & XMEAS 21             \\
Separator Cooling Water Outlet Temperature      & XMEAS 22             \\
\hline
\end{tabular}
\caption{Monitored variables of the TEP.}
\label{table:1}
\end{table}

\newpage
\vskip 0.2in
\bibliography{sample}

\end{document}